# LLM-guided headline rewriting for clickability enhancement without clickbait

Yehudit Aperstein[1, *], Linoy Halifa[1], Sagiv Bar[1], Alexander Apartsin[2]

[1]Intelligent Systems, Afeka Academic College of Engineering, Tel Aviv 69988, Israel
[2]School of Computer Science, Faculty of Sciences, HIT-Holon Institute of Technology, Holon 58102, Israel
[*]Corresponding Author: apersteiny@afeka.ac.il

## Abstract

Enhancing reader engagement while preserving informational fidelity is a central challenge in controllable text generation for news media. Optimizing news headlines for reader engagement is often conflated with clickbait, resulting in exaggerated or misleading phrasing that undermines editorial trust. We frame clickbait not as a separate stylistic category, but as an extreme outcome of disproportionate amplification of otherwise legitimate engagement cues. Based on this view, we formulate headline rewriting as a controllable generation problem, where specific engagement-oriented linguistic attributes are selectively strengthened under explicit constraints on semantic faithfulness and proportional emphasis.

We present a guided headline rewriting framework built on a large language model (LLM) that uses the Future Discriminators for Generation (FUDGE) paradigm for inference-time control. The LLM is steered by two auxiliary guide models: (1) a clickbait scoring model that provides negative guidance to suppress excessive stylistic amplification, and (2) an engagement-attribute model that provides positive guidance aligned with target clickability objectives. Both guides are trained on neutral headlines drawn from a curated real-world news corpus. At the same time, clickbait variants are generated synthetically by rewriting these original headlines using an LLM under controlled activation of predefined engagement tactics.

By adjusting guidance weights at inference time, the system generates headlines along a continuum from neutral paraphrases to more engaging yet editorially acceptable formulations. The proposed framework provides a principled approach for studying the trade-off between attractiveness, semantic preservation, and clickbait avoidance, and supports responsible LLM-based headline optimization in journalistic settings.

## 1. Introduction

Digital news headlines function as the primary entry point to journalistic content and strongly influence readers' decisions about what to consume. In contemporary, algorithmically mediated information environments, headlines must attract attention within a narrow cognitive window while remaining faithful to the substance of the underlying article. This creates a structural tension between engagement and editorial integrity. In practice, this tension has encouraged the use of clickbait-like strategies, which can increase short-term attention but are often associated with reader frustration, unmet expectations, and long-term erosion of trust in news outlets.

Existing research has largely addressed this problem through clickbait detection, classification, and filtering, implicitly treating it as a discrete, undesirable stylistic category. Headlines are typically labeled as clickbait or non-clickbait, and models are trained to identify lexical, syntactic, or pragmatic markers correlated with exaggeration or deception. While this work has produced useful datasets and benchmarks, it obscures an important distinction: many linguistic devices associated with clickbait, such as curiosity induction, emphasis framing, specificity manipulation, or interrogative constructions, are standard rhetorical tools in journalism. These mechanisms are not inherently misleading; they become problematic only when their use is amplified beyond editorially defensible bounds.

To make this distinction explicit, we define an engagement-attribute rubric that separates clickbait-oriented from clickability-oriented realizations of common headline mechanisms. The full comparison is presented later in Table 1 (Section 3.1.1), where these mechanisms are operationalized for synthetic data generation and controllable guidance. This comparison motivates a shift from binary classification to a continuous, controllable view of engagement. We therefore adopt a generative rather than purely discriminative perspective and posit two core assumptions. First, clickbait and clickability differ in degree rather than kind: clickbait arises when otherwise legitimate engagement mechanisms are amplified beyond proportionate or editorially acceptable levels. Second, engagement can be decomposed into controllable linguistic attributes, allowing individual mechanisms (e.g., curiosity, specificity, or framing) to be selectively strengthened or attenuated while maintaining explicit constraints on semantic fidelity and overall emphasis. From this perspective, the central challenge is not to eliminate engagement mechanisms, but to regulate their strength and interaction in a principled manner. This reframes headline optimization as a controlled text-generation problem, where engagement is modeled as a continuous, multidimensional space rather than a binary label. Such a formulation naturally motivates the use of large language models (LLMs) as flexible generators augmented with explicit control signals.

In this work, a controllable headline rewriting framework is introduced that operationalizes this view using a state-of-the-art LLM as the base generator, guided at inference time by auxiliary models. The LLM is steered by two complementary signals: a BERT-based clickbait scoring model that penalizes excessive stylistic amplification, and a BERT-based engagement attribute model that encourages specified clickability objectives. By integrating these signals within a guided decoding paradigm, the framework enables fine-grained control over both the degree and type of engagement expressed in rewritten headlines, supporting responsible engagement enhancement without semantic distortion or disproportionate emphasis.

The main contributions of this research are as follows:
- Headline rewriting is formulated as a controllable text generation task, in which clickbait is modeled as a graded consequence of disproportionate engagement-attribute activation rather than as a binary stylistic category.
- An interpretable space of engagement-driven rewriting directions is defined, enabling headline transformations to be represented along distinct linguistic dimensions that separate legitimate optimization from disproportionate clickbait-oriented variation.
- Dedicated positive and negative guidance models are trained to generate control signals for decoding, promoting target engagement attributes while suppressing excessive stylistic amplification.
- A dual-guidance inference-time rewriting framework is proposed, incorporating these learned signals into LLM-based headline generation.
- Guided decoding is shown to enable fine-grained control over engagement-related linguistic features while reducing uncontrolled stylistic drift.

The rest of the paper is organized as follows. Section 2 reviews related work on clickbait detection and controllable text generation. Section 3 presents the proposed methodology, including the engagement-attribute rubric, synthetic dataset construction, auxiliary guide models, and FUDGE-based controlled decoding. Section 4 describes the experimental setting and reports the results of the clickbait scoring model, the engagement attribute model, and the controlled rewriting framework. Section 5 concludes the paper and outlines directions for future work.

## 2. Related Work

### 2.1 Clickbait Detection

Research on clickbait detection has progressed from feature-engineered classifiers to transformer-based and more interpretable frameworks. Early studies by Biyani et al. [1] and Chakraborty et al. [2] established that clickbait can be detected through lexical, stylistic, and structural cues, while Potthast et al. [3] standardized the task through the Webis Clickbait Corpus 2017 and the Clickbait Challenge, enabling both binary and graded modeling of clickbait strength.

The field has since expanded in both scope and methodology. As reviewed in [4] clickbait research now spans not only detection algorithms, but also semantic manipulation, curiosity mechanisms, and credibility effects. Methodologically, recent work has shifted toward deep learning and pretrained language models. For example, [5] reports strong performance with a RoBERTa-Large-based detector and complements classification with LIME and SHAP for interpretability, while [6] extends the task to Urdu and highlights the value of sentence embeddings in low-resource settings.

Recent studies also explore more data-efficient and LLM-oriented paradigms. For instance, [7] shows that strong results can be achieved from headline text alone, even with very limited labeled data, whereas [8] finds that zero-shot and few-shot LLMs remain competitive and multilingual but still lag behind task-specific fine-tuned or prompt-tuned models. Together, these studies indicate that headline-level clickbait detection benefits from explicit supervision and task-adapted modeling rather than relying solely on general-purpose LLM inference.

At the same time, recent work suggests that clickbait should not be treated as a purely binary phenomenon. For example, [9] shows that engagement follows a curvilinear relation with headline concreteness, implying that headlines can be too vague or too explicit. Similarly, [10] demonstrates that explicit informativeness features improve both performance and interpretability. Finally, [11] moves beyond binary detection toward tactic-level explanation by separating clickbait detection from rhetorical strategy attribution. Collectively, this line of work motivates treating clickbait as a graded, explainable, and attribute-driven phenomenon, which directly aligns with the present paper's controllable rewriting perspective.

### 2.2 Controllable Text Generation

In parallel with advances in detection, a growing body of work has explored controllable text generation, aiming to steer language model outputs toward desired attributes such as style, sentiment, topic, or level of sensationalism. Early approaches emphasized training-time control. Conditional language models explicitly incorporate control variables during training, enabling the generator to condition its output on metadata or attribute labels. A prominent example is CTRL [12], a large-scale transformer trained with control codes that specify domains, styles, or other high-level properties. By learning these associations directly during pretraining, CTRL can generate text aligned with the provided control signal at inference.

Another training-time paradigm relies on reinforcement learning, where generation policies are optimized using reward functions that encode target attributes. While effective, such approaches are often computationally expensive and tightly coupled to specific control objectives.

More recently, research has increasingly focused on inference-time control methods that operate on a fixed, pretrained language model, avoiding full retraining. These approaches augment the decoding process with auxiliary models or signals that bias token selection. Plug-and-Play Language Models [13] exemplify this strategy by combining a frozen generator with lightweight attribute classifiers. During generations, gradients from the classifier modify the language model's hidden states at each step, enabling control over attributes such as topic or sentiment without altering the base model's parameters.

A related probabilistic method is GeDi [14] which uses smaller generative models as discriminators. At each decoding step, GeDi computes class-conditioned probabilities for the desired attribute and its complement, applying Bayes' rule to reweight the token distribution of a larger generator. Another decoding-time approach, DExperts [15], casts control as a product-of-experts formulation, combining an "expert" model trained on desirable text with an "anti-expert" trained on undesirable text, so that only tokens favored by the expert and disfavored by the anti-expert are promoted.

An effective inference-time control framework is introduced in FUDGE [16], where a discriminator predicts whether a partially generated sequence will satisfy a target attribute. During decoding, the base language model's next-token probabilities are reweighted by the discriminator's estimate, yielding a principled Bayesian factorization of $P$(text | attribute) at inference time. This enables control over attributes such as style or formality without modifying or fine-tuning the underlying generator.

These methods demonstrate that fine-grained controllability can be achieved by augmenting pretrained language models at inference. In particular, approaches such as FUDGE highlight how lightweight, modular discriminators can steer fluent generation toward desired attributes through probabilistically grounded mechanisms, making them well-suited for applications that require subtle control, such as generating curiosity-inducing yet non-clickbait headlines, while preserving linguistic quality.

# 3. Methodology

The proposed approach formulates headline rewriting as a controlled generation task in which a neutral source headline is transformed into a more engaging alternative while limiting excessive stylistic amplification and preserving semantic faithfulness. Figure 1 summarizes the general flow of the proposed rubric-based and FUDGE-guided headline rewriting approach. In the first stage, neutral source headlines and an engagement-attribute rubric are used to construct a synthetic training dataset. In the second stage, this dataset is used to train two auxiliary guide models: a clickbait scoring model and an engagement attribute model. In the final stage, both guides are integrated with the base language model during FUDGE-style decoding to steer headline rewriting toward engaging yet restrained outputs.

## 3.1 Engagement Attribute Rubric

We first define a structured rubric of engagement attributes commonly found in news headlines, including information-gap control, emphasis intensity, emotional framing, salience allocation, referential clarity, structural emphasis, and interrogative framing (Table 1). For each attribute, the rubric specifies paired interpretations corresponding to clickability-oriented and clickbait-oriented usage. These paired interpretations should be understood not as different surface forms, but as different uses of the same underlying rhetorical mechanism. For example, information-gap control may appear either as calibrated partial disclosure that invites interest while preserving the core claim, or as withholding of essential facts that manufactures unresolved curiosity. Similarly, emphasis intensity may range from proportional emphasis that reflects the actual significance of an event to exaggerated amplification that overstates its importance. Across the remaining attributes, the same principle applies: clickability reflects bounded, content-faithful use of engagement devices, whereas clickbait reflects disproportionate or misleading use.

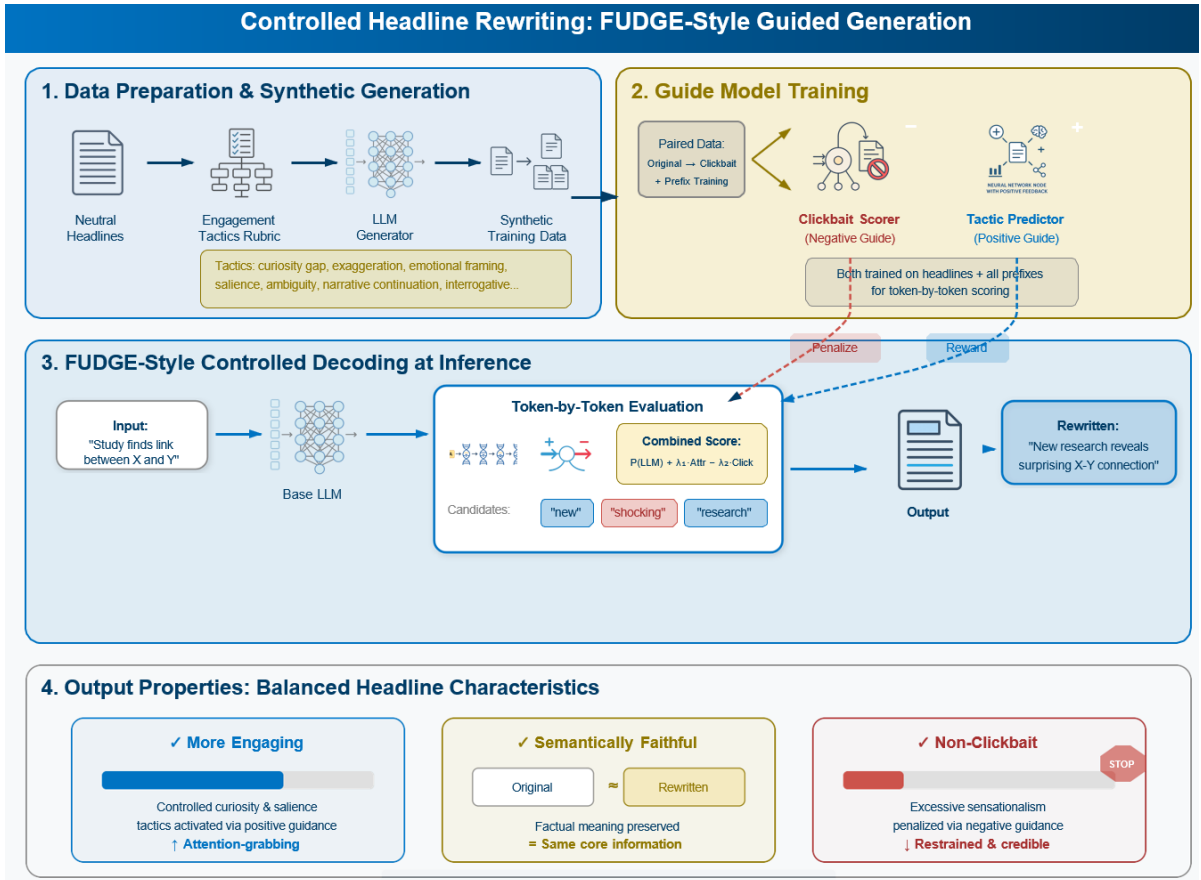

**Figure 1.** General flow of the proposed rubric-based and FUDGE-guided headline rewriting approach.

This rubric provides an explicit operationalization of engagement as a set of controllable linguistic dimensions rather than an undifferentiated stylistic label. Accordingly, the rubric is used both to define the attribute configurations employed in synthetic headline generation and to guide supervision of the auxiliary engagement model. It serves as the foundation for both data generation and model supervision.

| Attribute | Clickbait Interpretation | Clickability Interpretation |
| --- | --- | --- |
| Information gap control | **Curiosity gap** Deliberate withholding of essential facts to compel a click by creating unresolved curiosity | **Information gap calibration** Partial disclosure that invites interest while preserving the core factual claim |
| Emphasis intensity | **Exaggeration** Overstating importance, impact, or certainty beyond what is supported by the article | **Proportional emphasis scaling** Adjusting linguistic intensity to match the true significance of the event |
| Emotional framing | **Emotional trigger** Manipulative activation of fear, outrage, or excitement disproportionate to the facts | **Affective salience framing** Highlighting emotionally relevant aspects implied by the content without inflation |
| Salience allocation | **Sensationalism** Focusing on dramatic or marginal aspects at the expense of informational balance | **Salience-oriented framing** Foregrounding the most newsworthy element while maintaining contextual integrity |
| Structural emphasis | **Lists or superlatives** Use of absolute rankings or hyperbolic enumeration to imply exceptionalism | **Structured highlighting** Bounded use of ordering or comparison to organize key information |
| Referential clarity | **Ambiguous references** Vague entities or pronouns that obscure meaning to provoke curiosity | **Referential under-specification** Temporary abstraction that is resolved through reading without misleading |
| Reader addressing | **Direct appeals** Imperative or coercive language explicitly urging the reader to click | **Reader relevance cues** Implicit signaling of why the story matters to the reader |
| Narrative structure | **Unfinished narratives** Artificial cliffhangers that omit essential outcomes | **Narrative continuation cues** Indicating that further context follows while retaining essential information |
| Conceptual framing | **Unexpected associations** Forced or misleading linkage between unrelated concepts | **Cross-domain framing** Novel but factually grounded connections that increase interpretive interest |
| Interrogative form | **Provocative questions** Questions implying false premises or exaggerated stakes | **Interrogative framing** Questions used to frame inquiry or uncertainty without presupposition |

**Table 1**: Clickbait vs. Clickability Interpretations of Common Headline Attributes

## 3.2 Synthetic Dataset Generation with Controlled Clickbait Interpretation

Using the engagement rubric, we generate a synthetic corpus of clickbait headlines via attributed prompting of a large language model (GPT-4o-mini), starting from a curated collection of real, neutral news headlines drawn from the True.csv portion of the ISOT Fake News Dataset [17,18]. Because this subset was collected from Reuters.com, we treat these headlines as neutral, fact-oriented baseline formulations and do not generate additional neutral variants. Each source headline defines one data sample and is assigned a non-clickbait label with an all-zero engagement attribute vector.

For every source headline, we generate a single clickbait variant by prompting the model to rewrite it using a controlled subset of engagement attributes. Specifically, the prompt allows the activation of up to three clickbait-oriented engagement mechanisms (e.g., Curiosity, Exaggeration, Emotional Triggers), which are explicitly specified at generation time. The generation is constrained to exaggerate only the selected dimensions while preserving the original headline's underlying factual content and core semantic meaning.

As a result, each generated sample consists of a headline text, a binary clickbait label, and a sparse binary engagement-attribute vector indicating which attributes were exaggerated. This construction yields a dataset in which neutral headlines and clickbait headlines represent two controlled extremes of the engagement space. The resulting data provides clean supervision for training both the binary clickbait scoring model and the engagement attribute model, while maintaining semantic alignment with the original curated headlines.

Table 2 shows representative examples from the synthetic dataset construction process. Each example pairs a real, curated neutral headline with an explicitly specified clickbait tactic and the resulting generated headline. The table illustrates how a fixed prompt template, differing only in the activated tactic, yields controlled, interpretable stylistic amplification. These examples demonstrate that the dataset encodes clickbait behavior at the level of individual engagement mechanisms, supporting both binary clickbait supervision and attribute-level modeling.

| Neutral Source Headline | Attribute-Control Prompt (excerpt) | Generated Headline |
|---|---|---|
| *The White House budget chief expects delay in hitting debt limit* | Rewrite the headline using the **Curiosity Gap** tactic. Deliberately withhold essential details to create unresolved curiosity. Do not add new facts. | *The White House budget chief reveals shocking delay in debt limit* |
| *Lawsuit says North Carolina bathroom law still harmful* | Rewrite the headline using **Exaggeration**. Overstate the importance and impact of the decision beyond what is directly supported by the article. | *You won't believe the impact of North Carolina's bathroom law.* |
| *Switzerland urges voters to keep the fee for public broadcasters.* | Rewrite the headline using an **Emotional Trigger**. Activate concern or alarm disproportionate to the reported findings. | *What remains hidden about Switzerland's public broadcaster fee debate* |
| *No country for migrant stowaways caught on ferry between Ukraine and Turkey.* | Rewrite the headline using **Lists or Superlatives**. Imply exceptionalism through hyperbolic ranking or enumeration. | *The dramatic truth behind migrant stowaways caught on a ferry* |

**Table 2**: Examples of Attribute-Controlled Clickbait Generation via Explicit Tactics

## 3.3 Clickbait Scoring Model

The clickbait scoring model is designed to operate on partial generations rather than on complete headlines alone, as it is used to guide the base language model during inference-time decoding. In this setting, the model must evaluate stylistic amplification from incomplete text, since guidance is applied incrementally as tokens are generated. Training a scorer solely on finished headlines would therefore be insufficient, as it would provide no supervision for intermediate prefixes.

To address this requirement, each headline in the synthetic dataset is decomposed into all of its token-level prefixes. Given a headline represented as a token sequence, $w_1, \ldots, w_T$, we construct training inputs $w_1, \ldots, w_t \ for \ all \ t \in \{1, \ldots, T\}$. Each prefix inherits the clickbait label of the full headline, allowing the model to learn how clickbait-related signals accumulate progressively as the headline unfolds. To explicitly distinguish complete headlines from prefixes, a terminal period token is appended to the full headline during training, while prefixes are left unmarked. This simple convention enables the model to condition its predictions on whether the input represents a completed headline or an intermediate generation state.

The scoring model itself is implemented by fine-tuning a BERT encoder with a binary classification head. For a given partial headline, the model outputs a scalar clickbait score, interpreted as the predicted probability that the current prefix belongs to the clickbait class. During guided decoding, this score is used as a negative control signal to penalize candidate continuations that increase clickbait likelihood. Because very short prefixes provide limited semantic evidence and are unreliable indicators of stylistic intent, we employ a length-aware weighted loss function during training. Loss contributions from shorter prefixes are down-weighted relative to those from longer prefixes and complete headlines, encouraging the model to focus on informative contexts while remaining robust to partial inputs. This training strategy yields a clickbait scorer that produces stable, prefix-consistent signals suitable for guiding controlled headline rewriting at inference time.

## 3.4 Engagement Attribute Model

In parallel with the clickbait scoring model, we train an engagement attribute model to predict the presence, and when applicable, the strength, of specific engagement mechanisms defined in the rubric. The model is trained on the same synthetic dataset and leverages structured supervision from explicit attribute annotations. Whereas the clickbait scorer provides a global assessment of disproportionate amplification, the engagement model captures fine-grained, attribute-specific signals corresponding to distinct rhetorical tactics.

The engagement model is implemented by fine-tuning a pretrained BERT encoder with a multi-class, multi-label prediction head. Each attribute is modeled as an independent prediction dimension, allowing the model to estimate whether a given engagement mechanism is present in the input. This design supports simultaneous prediction of multiple attributes and enables selective activation during inference-time guidance. As the model is used to guide generation incrementally, training is performed on both complete headlines and all token-level prefixes derived from them.

To support prefix-level operation, each headline is decomposed into its prefixes, and all prefixes inherit the attribute annotations of the full headline. Complete headlines are marked with a terminal punctuation token, while prefixes remain unmarked, enabling the model to distinguish between finished and partial inputs. Training employs a length-aware weighted loss function, in which each example's contribution is scaled by its prefix length. Very short prefixes, which provide weak evidence for attribute realization, are assigned lower weights, while longer prefixes and complete headlines are emphasized.

## 3.5 FUDGE-Based Controlled Decoding

To enable fine-grained control over headline rewriting, we adopt a FUDGE-based (Future Discriminators for Generation) inference-time control framework, in which generation from a base large language model is steered by auxiliary scoring models operating on partial outputs. In this setting, guidance is applied incrementally during decoding by modifying token-selection probabilities based on predicted properties of the generated prefix.

Let $x_{1:t}$ denote the current generation prefix and **y** a candidate next token. At each decoding step, the base LLM proposes candidate's continuations, which are then re-weighted using guidance scores derived from the auxiliary models evaluated on the extended prefix: $x_{1:t} \oplus y$. The resulting guidance scores are then used to adjust the relative probabilities of candidate next tokens before sampling or selection.

### 3.5.1 Negative Guidance via Clickbait Scoring

The clickbait scoring model provides a negative control signal intended to suppress disproportionate stylistic amplification. For each candidate's continuation, the scorer outputs a scalar score $S_{cb}(x_{1:t} \oplus y)$ representing the likelihood of clickbait. This score is incorporated as a penalty term with an associated weight $\lambda_{neg}$, reducing the probability of tokens that increase the predicted clickbait score. Intuitively, this signal acts as a soft constraint that discourages the generator from entering clickbait-prone regions of the engagement space without imposing hard rejection.

### 3.5.2 Positive Guidance via Engagement Attribute Signals

In parallel, the engagement attribute model provides positive control signals aligned with target clickability objectives. For each attribute k, the model produces a normalized score $S_k(x_{1:t} \oplus y)$ indicating the strength of that engagement mechanism. Desired attributes are assigned positive weights, while discouraged attributes are assigned negative weights. The overall positive guidance signal is computed as a weighted sum

$$S_{pos}(x_{1:t} \oplus y) = \sum_k \omega_k S_k(x_{1:t} \oplus y), \quad (1)$$

where $w_k \in \{-1, 0, 1\}$ encodes whether an attribute is encouraged, ignored, or actively suppressed. This formulation allows explicit specification of which engagement mechanisms should be strengthened and which should be avoided.

### 3.5.3 Combined Guidance Objective

The final token selection score combines the base LLM log-probability with both guidance terms:

$$\log p(y|x_{1:t}) \propto \log p_{LLM}(y|x_{1:t}) + \lambda_{pos} S_{pos}(x_{1:t} \oplus y) - \lambda_{neg} S_{cb}(x_{1:t} \oplus y) \quad (2)$$

By adjusting the guidance weights $\lambda_{neg} \geq 0$ and $\lambda_{pos} \geq 0$, the system can smoothly trade off engagement enhancement against editorial restraint.

This FUDGE-based formulation enables the base LLM to navigate the engagement space in a controlled, interpretable manner. Positive signals encourage the realization of selected clickability attributes, while negative signals continuously penalize excessive amplification. Together, these complementary controls enable responsible headline rewriting that enhances reader appeal while preserving semantic fidelity and avoiding clickbait extremes.

## 4. Experiments and Results

### 4.1 Dataset Overview

Figure 2 summarizes the composition of the synthetic dataset by engagement intensity according to the methodology presented in Section 3.1.2. Neutral headlines, drawn directly from the curated source corpus, contain no activated engagement tactics, while clickbait variants are generated by activating one, two, or three tactics per headline. The dataset is approximately balanced across clickbait groups, enabling the scoring model to learn a graded notion of clickbait strength while preserving a clear distinction between neutral and increasingly amplified formulations.

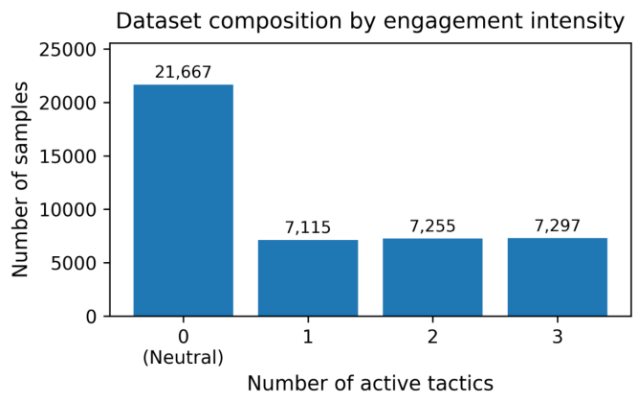

**Figure 2.** Dataset composition by engagement intensity. Neutral headlines contain 0 active tactics, whereas clickbait variants are generated with 1, 2, or 3 active tactics.

### 4.2 Clickbait Scoring Performance

For training and evaluation the clickbait scoring model, the data are split at the source headline level prior to synthetic generation to prevent leakage across prefixes. The curated set of real neutral headlines is first divided into 80% for training/validation and 20% for testing. Synthetic clickbait variants and their corresponding prefixes are then generated separately within each split. Model fine-tuning is conducted up to 3 epochs, with early stopping based on validation loss to prevent overfitting. Optimization uses the AdamW optimizer with a learning rate of $2\times10^{-5}$ and linear decay thereafter. Mini-batches are constructed by sampling prefixes of varying lengths to ensure exposure to both partial and near-complete contexts within each batch.

On the held-out synthetic test set, the clickbait scorer achieved AUROC values above 0.99, indicating near-perfect discrimination between neutral and clickbait headlines, which is expected given the controlled synthetic setup and the explicit rubric used to generate clickbait variants.

### 4.3 Engagement Attribute Model Performance

For training and evaluation the engagement attribute model, the dataset was split at the source headline level into training, validation, and test sets, with 80% allocated to training (65%) and validation (15%) and 20% to the test set to prevent leakage across prefixes. Fine-tuning is performed trained for up to 3 epochs, with early stopping based on validation loss. Optimization uses the AdamW optimizer with a learning rate of $2\times10^{-5}$. During guided decoding, the engagement attribute model supplies positive, attribute-specific guidance signals that encourage the base LLM to realize selected engagement mechanisms. In contrast, the clickbait scoring model provides a complementary negative signal that constrains excessive amplification.

Figure 3 summarizes the behavior of the engagement attribute model through two complementary visualizations. Although the model is trained in a multi-label setting, we report per-tactic results on single-tactic examples for interpretability. This isolates each engagement mechanism and makes it possible to analyze tactic-specific recognition quality and structured confusions without the ambiguity introduced by overlapping labels. The bar plot shows differences in how reliably various engagement tactics are detected, with some attributes, such as Lists/Superlatives and Curiosity Gap, appearing more consistently recognizable. In contrast, others, including Sensationalism and Ambiguous References, are more difficult to identify.

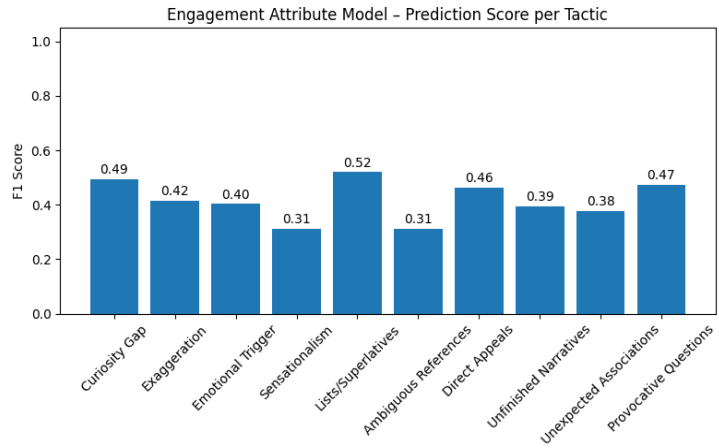

**Figure 3**: Normalized prediction score per tactics

The confusion matrix (Figure 4) provides a more detailed view of the model's predictions for single-attribute examples. A clear diagonal pattern indicates that the model often identifies the correct dominant tactic, while structured off-diagonal errors reveal confusion between semantically related attributes, such as emotional triggers and sensational phrasing. The results suggest that the model captures distinct engagement patterns while reflecting the natural overlap among persuasive strategies, providing informative and interpretable signals to guide controlled headline generation.

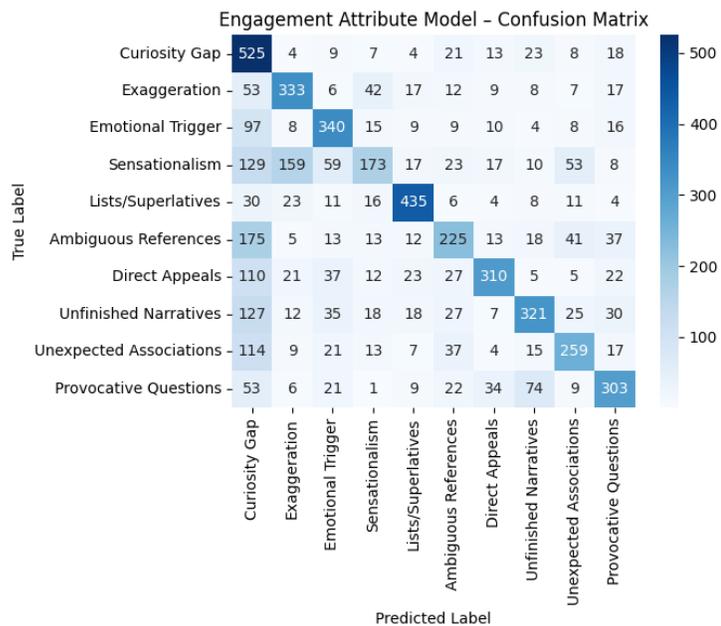

**Figure 4**: Confusion matrix for single-tactic examples

## 4.4 Controlled rewriting results

Table 3 presents qualitative examples of headline rewrites targeting different engagement attributes. The neutral headlines report events factually and descriptively without rhetorical framing. When rewritten without clickbait control ($\lambda_{neg}= 0$, $\lambda_{pos}= 0.5$), the generated headlines tend to increase engagement by employing stronger rhetorical strategies, such as curiosity gaps, exaggeration, metaphorical associations, emotional triggers, and provocative questions. These formulations often use more dramatic or sensational language to increase reader interest. When clickbait control is applied ($\lambda_{pos}= 0.5$, $\lambda_{neg}= 1$), the headlines remain engaging and continue to reflect the intended rhetorical tactic, but the phrasing becomes more restrained and informative. These examples illustrate that the proposed framework can generate headlines that express specific engagement attributes while moderating excessive sensationalism through the clickbait control mechanism.

| Target engagement attribute | Neutral headline | Rewriting (no clickbait control, $\lambda_{neg}= 0$, $\lambda_{pos} = 0.5$) | Rewriting (clickbait-controlled, $\lambda_{pos}= 0.5$, $\lambda_{neg}= 1$) |
|---|---|---|---|
| Curiosity gap | *European football: Harry Kane breaks personal record, Dortmund suffer loss* | *What is behind Harry Kane's remarkable milestone in European football, as Dortmund's defeat adds to the drama?* | *European Football: Can Harry Kane's Unstoppable Form Breed a New Era of Success?* |
| Exaggeration | *Venezuelan crisis spawns boom in gambling.* | *Venezuela's Desperate Situation Ignites a Frenzy of High-Stakes Bets.* | *Venezuela's Desperate Struggle Highlights Troubled Citizens' Frenzied Quest for Luck.* |
| Unexpected associations | *Arctic drilling hits speed bumps in U.S. tax bill* | *Frozen ambitions stall in the unexpectedly icy waters of fiscal legislation.* | *Frozen ambitions stall in Washington's fiscal freeze.* |
| Curiosity Gap + Emotional Trigger | *Apple case exposes ongoing government rift over encryption policy* | *What is Behind the Apple Case: A Deepening Divide on Encryption Policy.* | *Behind Closed Doors: The Apple Case Reveals a Deepening Divide Over Encryption Secrets.* |
| Curiosity Gap + Provocative Questions | *Lockheed Martin wins $450 million Pentagon contract, statement says.* | *What is Behind the Pentagon's $450 Million Investment in Lockheed Martin?* | *Lockheed Martin Lands Lucrative Pentagon Deal: But at What Cost to Taxpayers?* |

**Table 3**: Examples of headline rewrites targeting specific engagement attributes

# 5. Conclusion and Future Work

This work demonstrates that engagement-oriented headline rewriting can be explicitly controlled to avoid degenerating into clickbait. By introducing separate positive and negative control coefficients, the proposed approach disentangles *engagement enhancement* from *manipulative amplification*. Empirical examples show that unconstrained rewriting ($\lambda_{neg}= 0$) reliably increases attention-grabbing cues but frequently introduces exaggeration, implied secrecy, or emotional distortion. In contrast, controlled rewriting with simultaneous engagement promotion and clickbait suppression produces headlines that are more restrained, more semantically faithful to the source content, and better aligned with the intended balance between attractiveness and proportionality. Moreover, clickbait suppression ($\lambda_{neg} > 0$) yields headlines that remain semantically faithful to the source content while exhibiting measurable gains in engagement salience.

The results support the central claim that clickbait is not an inevitable byproduct of engagement optimization. Instead, it emerges when negative constraints are absent. Framing headline generation as a constrained optimization problem provides a practical, interpretable mechanism for navigating the trade-off between informativeness and appeal, with direct relevance to responsible media generation, recommender systems, and editorial AI tools.

Several directions naturally extend this study. First, the control framework can be expanded from scalar coefficients to tactic-specific penalties, allowing fine-grained suppression of individual clickbait strategies such as urgency inflation, emotional manipulation, or curiosity gaps. Second, future work should evaluate the approach at scale using human judgments and real-world engagement metrics, enabling calibration of $\lambda_{pos}$ and $\lambda_{neg}$ against editorial or platform-specific standards. Third, integrating the control mechanism into multi-step or reinforcement-based generation loops may further improve robustness under tactic composition. Finally, extending the framework beyond headlines to longer-form content such as summaries or social media posts would test its generality and applicability to broader content moderation and responsible text-generation settings.

## Data Availability

The dataset and code are publicly available at: https://github.com/LinoyHalifa/Clickbait-Research

## Funding

This research received no specific grant from any funding agency in the public, commercial, or not-for-profit sectors.

## References

1. Biyani, P., Tsioutsiouliklis, K., & Blackmer, J. (2016). "8 Amazing Secrets for Getting More Clicks": Detecting Clickbaits in News Streams Using Article Informality. Proceedings of the 30th AAAI Conference on Artificial Intelligence (AAAI-16), 94–102.
2. Chakraborty, A., Paranjape, B., Kakarla, S., & Ganguly, N. (2016). Stop Clickbait: Detecting and Preventing Clickbaits in Online News Media. In Proceedings of IEEE/ACM ASONAM 2016.
3. Potthast, M., Gollub, T., Hagen, M., & Stein, B. (2018). The Clickbait Challenge 2017: Towards a Regression Model for Clickbait Strength. arXiv preprint arXiv:1812.10847.
4. Jácobo-Morales, D., & Marino-Jiménez, M. (2024). Clickbait: Research, challenges and opportunities–A systematic literature review. *Online Journal of Communication and Media Technologies*, *14*(4), e202458.
5. Alarfaj, F. K., Muqadas, A., Khan, H. U., & Naz, A. (2025). Clickbait detection in news headlines using RoBERTa-Large language model and deep embeddings. *Scientific Reports*.
6. Muqadas, A., Khan, H. U., Ramzan, M., Naz, A., Alsahfi, T., & Daud, A. (2025). Deep learning and sentence embeddings for detection of clickbait news from online content. *Scientific Reports*, *15*(1), 13251.

7. Wang, Y., Zhu, Y., Li, Y., Qiang, J., Yuan, Y., & Wu, X. (2024). Clickbait detection via prompt-tuning with titles only. *IEEE Transactions on Emerging Topics in Computational Intelligence*, *9*(1), 695-705.
8. Wang, H., Zhu, Y., Wang, Y., Li, Y., Yuan, Y., & Qiang, J. (2025, July). Clickbait detection via large language models. In *International Conference on Intelligent Computing* (pp. 462-474). Singapore: Springer Nature Singapore.
9. Aubin Le Quéré, M., & Matias, J. N. (2025). When curiosity gaps backfire: effects of headline concreteness on information selection decisions. *Scientific Reports*, *15*(1), 994.
10. Michaluk, W., Urban, T., Kubita, M., Kuntur, S., & Wroblewska, A. (2026). Click it or Leave it: Detecting and Spoiling Clickbait with Informativeness Measures and Large Language Models. *arXiv preprint arXiv:2602.18171*.
11. Nofar, L., Portal, T., Elbaz, A., Apartsin, A., & Aperstein, Y. (2025). An Interpretable Benchmark for Clickbait Detection and Tactic Attribution. arXiv preprint arXiv:2509.10937.
12. Keskar, N. S., McCann, B., Varshney, L. R., Xiong, C., & Socher, R. (2019). CTRL: A Conditional Transformer Language Model for Controllable Generation. arXiv:1909.05858.
13. Dathathri S, Madotto A, Lan J, Hung J, Frank E, Molino P, Yosinski J, Liu R. Plug and play language models: A simple approach to controlled text generation. arXiv preprint arXiv:1912.02164. 2019 Dec 4.
14. Krause, B., Gotmare, A. D., McCann, B., Keskar, N. S., Joty, S., Socher, R., & Rajani, N. F. (2021). GeDi: Generative Discriminator Guided Sequence Generation. Findings of EMNLP 2021, 4929–4952.
15. Liu, A., Sap, M., Lu, X., Swayamdipta, S., Bhagavatula, C., Smith, N. A., & Choi, Y. (2021). DExperts: Decoding-Time Controlled Text Generation with Experts and Anti-Experts. In Proceedings of ACL 2021.
16. Yang, K., & Klein, D. (2021). FUDGE: Controlled Text Generation with Future Discriminators. In Proceedings of NAACL 2021, 3511–3535.
17. Ahmed, H., Traore, I., & Saad, S. (2018). Detecting opinion spams and fake news using text classification. Security and Privacy, 1(1), e9.
18. Ahmed, H., Traore, I., & Saad, S. (2017, October). Detection of online fake news using n-gram analysis and machine learning techniques. In International conference on intelligent, secure, and dependable systems in distributed and cloud environments (pp. 127-138). Cham: Springer International Publishing.